%% file: root.tex
\documentclass[letterpaper, 10 pt, journal, twoside]{ieeetran}

\IEEEoverridecommandlockouts                              %

\usepackage[letterpaper, margin=59pt, top=61pt, bottom=43pt, left=48pt, right=48pt, includehead]{geometry}

\usepackage{cite}  %
\usepackage{gensymb}  %
\usepackage{array}  %
\usepackage{booktabs}  %
\usepackage[hidelinks]{hyperref}  %
\usepackage{subcaption}  %
\usepackage{siunitx}  %
\usepackage{amssymb}  %
\usepackage[normalem]{ulem} %
\usepackage{threeparttable} %
\usepackage{makecell} %

\newcommand{\legendbox}[1]{%
  \textcolor{#1}{\rule{6pt}{6pt}}%
}

\usepackage[acronym]{glossaries}
\newacronym{rl}{RL}{reinforcement learning}
\newacronym{il}{IL}{imitation learning}
\newacronym{ras}{RAS}{Robot-Assisted Surgery}
\newacronym{iou}{IoU}{Intersection over Union}
\newacronym{lc}{LC}{laparoscopic cholecystectomy}
\newacronym{rcm}{RCM}{remote center of motion}

\usepackage{xargs}                      %
\usepackage[pdftex,dvipsnames]{xcolor}  %
\usepackage[colorinlistoftodos,prependcaption,textsize=tiny]{todonotes}
\newcommandx{\question}[2][1=]{\todo[linecolor=red,backgroundcolor=red!25,bordercolor=red,#1]{#2}}
\newcommandx{\toadd}[2][1=]{\todo[linecolor=blue,backgroundcolor=blue!25,bordercolor=blue,#1]{#2}}
\newcommandx{\notes}[2][1=]{\todo[linecolor=OliveGreen,backgroundcolor=OliveGreen!25,bordercolor=OliveGreen,#1]{#2}}
\newcommandx{\improvement}[2][1=]{\todo[linecolor=Plum,backgroundcolor=Plum!25,bordercolor=Plum,#1]{#2}}

\definecolor{livercolor}{HTML}{CD9A8F}
\definecolor{reallivercolor}{HTML}{FF311A}
\definecolor{tubescolor}{HTML}{DA9341}
\definecolor{gallbladdercolor}{HTML}{E6D48A}
\definecolor{fattycolor}{HTML}{E6E38A}
\definecolor{realfattycolor}{HTML}{F3EC18}
\definecolor{targetgreen}{HTML}{3DCD6D}
\definecolor{targetred}{HTML}{cc0000}

\usepackage{xspace}

\usepackage{graphicx}
\usepackage{tikzexternal}
\tikzsetexternalprefix{tikz/}

\begin{document}

\title{%
Point Cloud Segmentation for Autonomous Clip Positioning in Laparoscopic Cholecystectomy on a Phantom
}

\author{Bal\'{a}zs Gyenes$^{1,2}$, Nikolai Franke$^{1}$, Paul Maria Scheikl$^{3}$, Pit Henrich$^{3}$, Rayan Younis$^{4}$,\\
Gerhard Neumann$^{1}$, Martin Wagner$^{4}$, Franziska Mathis-Ullrich$^{3}$%
\thanks{Manuscript received: January, 27, 2025; Revised May, 15, 2025; Accepted June, 25, 2025.}%
\thanks{This paper was recommended for publication by Editor Pietro Valdastri upon evaluation of the Associate Editor and Reviewers' comments. %
This work was supported by the Ministerium für Wirtschaft, Arbeit und Wohnungsbau, Baden-Württemberg, Germany, the Helmholtz Association under the joint research school ``HIDSS4Health – Helmholtz Information and Data Science School for Health”, and by the German Research Foundation (Deutsche Forschungsgemeinschaft) as part of Germany’s Excellence Strategy – EXC 2050/1 – Project ID 390696704 – Cluster of Excellence “Centre for Tactile Internet with Human-in-the-Loop” (CeTI) of Technische Universität Dresden.} %
\thanks{$^{1}$Institute for Anthropomatics \& Robotics - Karlsruhe Institute of Technology, Germany}%
\thanks{$^{2}$HIDSS4Health - Helmholtz Information and Data Science School for Health, Karlsruhe/Heidelberg, Germany}%
\thanks{$^{3}$Department of Artificial Intelligence in Biomedical Engineering, Friedrich-Alexander-University, Erlangen-Nuremberg, Germany}%
\thanks{$^{4}$University Hospital Carl Gustav Carus and Centre for Tactile Internet with Human-in-the-loop (CeTI), Dresden University of Technology, Germany}%
\thanks{Digital Object Identifier (DOI): see top of this page.}%
}

\markboth{IEEE Robotics and Automation Letters. Preprint Version. Accepted June, 2025}
{Gyenes \MakeLowercase{\textit{et al.}}: Point Cloud Segmentation for Autonomous Clip Positioning}

\maketitle

\begin{abstract}

High-risk applications in robotics, such as robot-assisted surgery, present unique challenges.
These systems must be both highly precise and interpretable in order to be deployed in environments with very low tolerance for error or unsafe exploration.
We present the first robotic system to demonstrate autonomous clip positioning on a physical phantom in laparoscopic surgery, one of the most common interventions in general surgery.
After segmentation of a colorless point cloud from a single camera, target positions for the clips are extracted using spline interpolation, and can then be adjusted by the human operator.
The segmentation model is trained on only 60 hand-labeled real point clouds, reflecting data scarcity in the surgical domain.
We overcome this with a combination of pre-training on 128,000 synthetic point clouds and two novel data augmentation techniques.
The motion of the end-effector to each target is visualized for the operator, satisfying the unique motion constraints of minimally-invasive surgery while ensuring that the robot's actions are verifiable and interpretable.
In real robot experiments, our system localizes targets with the required precision of 0.75~mm at a 95\% success rate and executes autonomous clip positioning with a 100\% success rate.
We provide insights that are applicable to many other surgical and non-surgical tasks that require identifying and navigating to a precise target.
Our source code is available at \href{https://github.com/balazsgyenes/kirurc}{https://github.com/balazsgyenes/kirurc}.

\end{abstract}


\section{INTRODUCTION}
\label{sec:introduction}

\begin{figure}[tb]
    \centering
    \includegraphics[trim={25cm 12.7cm 24.7cm 14cm},clip,width=\linewidth]{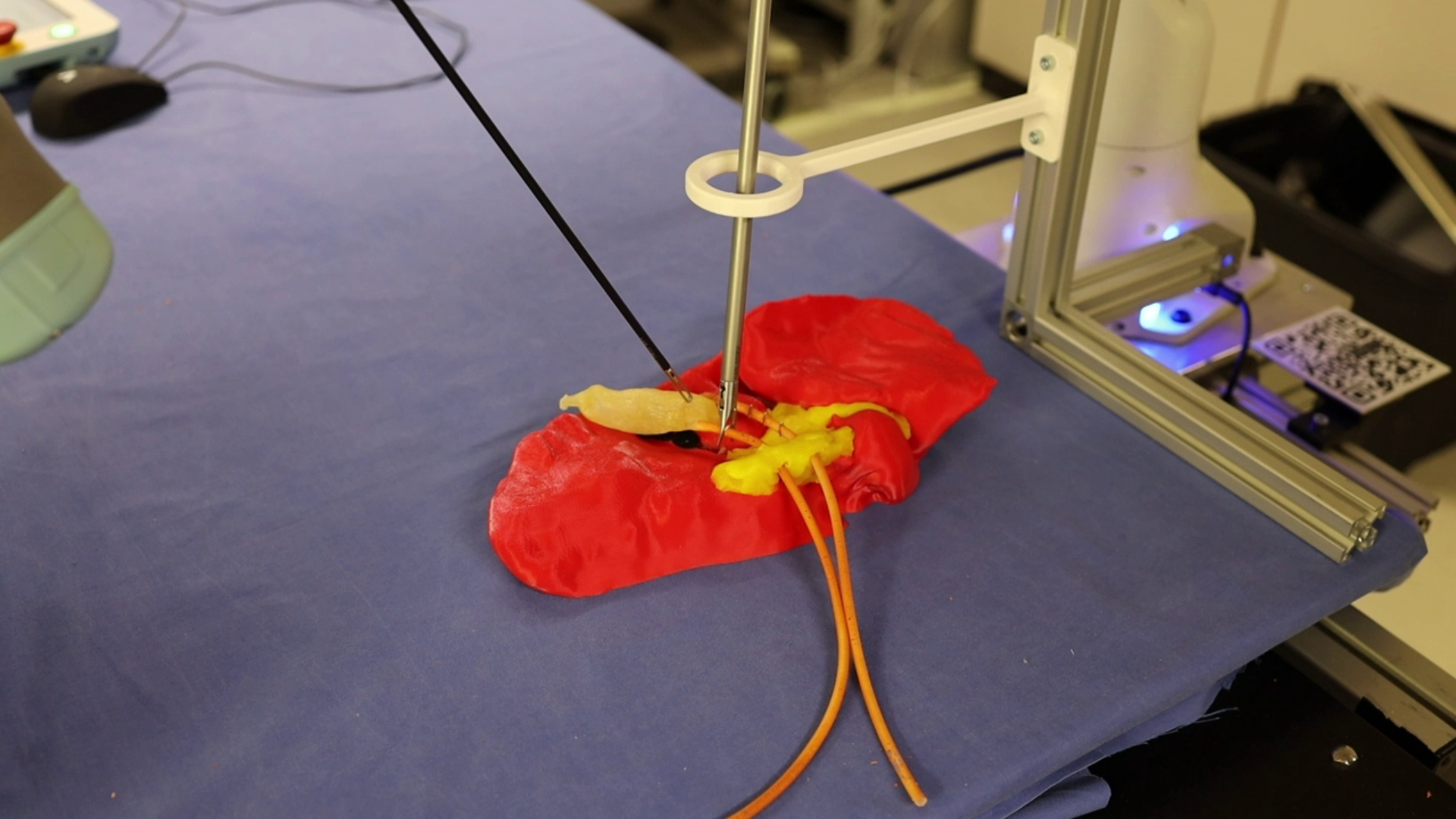}
    \caption{
    We present a robotic system for autonomous clip positioning in laparoscopic cholecystectomy.
    At this stage of the surgery, the cystic duct and cystic artery~(\legendbox{tubescolor}) have been dissected from the surrounding fatty tissue~(\legendbox{realfattycolor}) and are ready for ligation.
    Ligation is performed using clips applied by a laparoscopic clip applier (right tool).
    Next, the cystic duct and artery are cut to allow the gallbladder~(\legendbox{gallbladdercolor}) to be resected from its bed in the liver~(\legendbox{reallivercolor}).
    Gripper (left tool) provides tension.
    }
    \label{fig:real_setup}
    \vspace{-0.5cm}
\end{figure}
\glsreset{lc}

\IEEEPARstart{R}{ecent} years have seen many machine learning applications in robotics, especially with large generative models.
Although these systems can generalize across many simple tasks, high-risk applications require movements that are precise, yet interpretable and predictable, two characteristics not typical of large generative models.
One such high-risk application is \gls*{ras}, where (semi-)~autonomous policies have the potential to address the shortage of skilled personnel.
Semi-autonomous \gls*{ras} systems must be interpretable, since surgeons are unlikely to trust a system they cannot anticipate, and also extremely precise due to the small scale of relevant anatomical structures.
Furthermore, data scarcity is a wide-spread problem in medical contexts, so algorithms must generalize to unseen anatomies.
\Gls*{lc} is one of the most common interventions in general surgery~\cite{shafferEpidemiologyGallbladderStone2006}, so it provides a good case study for challenges that arise in \gls*{ras} and in other high-risk robotic applications.
During the clipping phase of \gls*{lc}, 3 clips are placed on the cystic artery and the cystic duct each. %
Accurately positioning the clips is a challenging task and a critical step for a successful surgery.

We present an autonomous learning-based robotic system to carry out clip positioning in laparoscopic surgery, demonstrating our method in the clipping phase of \gls*{lc} on a realistic 3D-printed phantom (see \autoref{fig:real_setup}).
In our phantom, the cystic artery and cystic duct are \SI{5}{\mm} in diameter, but each clip is only \SI{6.5}{\mm} wide, such that it must be placed with sub-millimeter precision.
Generalist end-to-end approaches, such as \gls*{rl}, \gls*{il}, or end-to-end prediction of the target positions, lack the required precision and may also not be interpretable.
To achieve both these requirements, we instead opt for a point cloud-based vision system that segments target regions and predicts targets in the form of B-splines.
A motion planner can then generate a complete trajectory to each target point in an interpretable way.
This corresponds to task-level autonomy as defined by Haidegger et al.~\cite{haideggerAutonomySurgicalRobots2019}, as the system independently generates plans to perform clip positioning, with human input only required to approve the plan and optionally adjust it.

The majority of image-based \gls*{ras} systems that use 3D information treat the depth map as an additional channel to the RGB image (RGBD).
Due to hardware limitations and the homogeneous and specular environment, the depth estimate is often noisy or inaccurate, so the RGBD image is rarely converted to a point cloud.
However, point clouds have several advantages over RGBD images, despite nominally containing the same information.
Point cloud models are more robust to changes in camera perspective~\cite{PCWM}, and tend to exhibit better sim-to-real transferability because point clouds focus on geometry rather than texture~\cite{qinDexPointGeneralizablePoint2023}.
This allows us to mitigate the data scarcity by pre-training on a large amount of synthetic data generated in simulation and fine-tuning using a small, manually labeled, real-world dataset.
Approaches that rely on RGB data require accurately rendering the colors and textures of the real scene, which is notoriously difficult.

In summary, we present the first system to demonstrate autonomous clip positioning in laparoscopic surgery on a physical phantom.
Rather than regressing 3D target positions directly, our key insight is to segment the target regions, skeletonize them, and fit them to B-splines, leveraging the accuracy of the 3D camera.
This results in a system which is precise, while remaining interpretable and human-correctable.
A human operator is able to visually validate or modify the target positions and, because the trajectory is computed in advance, they are able to confirm its safety before execution.
Despite using only 60 hand-labeled data points for training, reflecting the data scarcity in the surgical domain, we demonstrate a 95\% success rate at localizing targets within a tolerance of \SI{0.75}{\mm}.
To achieve this, we pretrain the segmentation model on a large synthetic dataset and leverage a set of straightforward but essential algorithmic contributions:
i) We demonstrate that pre-training with extra segmentation classes that are not present in the fine-tuning data help the model learn the semantic structure of the scene, but only if the extra classes are merged in \textit{logit space} rather than probability space.
ii) We introduce two new forms of point cloud data augmentation: \texttt{VariableJitter}, allowing random jitter to be applied without introducing a distribution gap; and \texttt{RandomPatches}, which provides robustness against missing points due to occlusions.
We validate our design choices post-hoc on a disjoint, real-world test set that was unavailable at training time.

\section{RELATED WORK}

\paragraph{Deep Learning for Segmentation \& Control in RAS}
Point clouds are under-explored in \gls*{ras}, partly because of the limited accuracy of 3D cameras with a form factor that can fit through a conventional \qtyrange[range-units=single,range-phrase=-]{10}{12}{\mm} trocar.
However, a number of works use RGB or RGBD information to identify a structure of interest and control an end-effector to interact with it.
Tanzi et al.~\cite{tanziRealtimeDeepLearning2021} segment an endoscopic image to align it to an ad-hoc 3D virtual model, achieving a system with an error of \SI{4.1}{\mm}.
Attanasio et al.~\cite{attanasioAutonomousTissueRetraction2020, attanasioComparativeStudySpatioTemporal2021} demonstrate tissue segmentation in abdominal surgery using RGBD images from a stereo endoscope.
They construct a custom step-based planning algorithm to grasp and retract a tissue flap.
Qi et al.~\cite{qiAutomaticPathPlanning2022} use a 3D U-Net architecture to segment preoperative CT images, after which the key points are detected.
A surgical path is pre-planned using a hand-crafted algorithm.
Schwaner et al.~\cite{schwanerAutonomousNeedleManipulation2021} learn a policy for autonomous surgical suturing from demonstrations and achieve a mean needle insertion error of \SI{3.8}{mm}.
Movements are encoded as Dynamic Movement Primitives, which offers a degree of interpretability compared to step-based inference.
Most similarly to our work, Lu et al.~\cite{luLearningDrivenFrameworkSpatial2020} develop a system to autonomously grasp a suture thread after identifying its 3D structure.
They fine-tune a model pre-trained on diverse data and achieve an average error between \SI{2.4}{mm} and \SI{4.5}{mm}.

\paragraph{Point Clouds in RAS}
A limited number of prior works operate directly with point clouds in the context of \gls*{ras}.
There are several examples of vision systems that use point cloud features to estimate 3D state, such as point-wise displacements of facial bone models for the planning of corrective jaw surgery~\cite{xiaoEstimatingReferenceBony2021} or estimating the 6D pose of a needle~\cite{zhou6DOFNeedlePose2019}.
Li et al.~\cite{liSuPerSurgicalPerception2020} create a stereo vision-based perception system that estimates the 3D state of the environment as surfels, which can easily be converted to point clouds.
Thach et al.~\cite{thachLearningVisualShape2022} learn a servoing policy from demonstrations using an observed point cloud and a goal point cloud as inputs, but  only achieve an accuracy of \SI{\sim 40}{\cm}.
The STAR system~\cite{saeidiAutonomousRoboticLaparoscopic2022} autonomously performs intestinal anastomosis in vivo on porcine models, relying on added visual markers and several task-specific algorithms and datasets, including 9294 labeled data points for learning tissue motion tracking.

\paragraph{Point Cloud Segmentation}
Point cloud segmentation is a well-studied task going back to the advent of PointNet~\cite{qiPointNetDeepLearning2017}.
Since then, a number of architectures have been proposed for segmentation as well as classification of point clouds.
These architectures make use of several paradigms, including graph neural networks~\cite{zhaoPointTransformer2021}, graph convolutions~\cite{wangDynamicGraphCNN2019}, and transformers~\cite{pangMaskedAutoencodersPoint2022,yuPointBERTPreTraining3D2022}.
PointTransformer~\cite{zhaoPointTransformer2021} is a graph-based model that makes use of a sequence of downsampling and upsampling layers, and achieves state of the art performance on segmentation of standard point cloud datasets.

\section{METHODS}
\label{sec:method}

\begin{figure*}[t]
    \begin{center}    
    \resizebox{0.8\textwidth}{!}{\input{figures/overview}}
    \end{center}
    
    \vspace{-0.3cm}
    \caption{
        Schematic of inference pipeline, showing how the input point cloud is segmented to isolate the target regions.
        Liver~(\legendbox{livercolor}), cystic duct and artery~(\legendbox{tubescolor}), gallbladder~(\legendbox{gallbladdercolor}), and fatty tissue~(\legendbox{fattycolor}) are colored as in the real scene, while target regions~(\legendbox{targetgreen}) are bright green for clarity.
        To increase robustness against outliers in the segmentation, each target region is then skeletonized and fitted to a B-spline.
        To ensure safe execution and accommodate a surgeon's preference, the human operator is able to customize the target position (\legendbox{targetred}) along the interpolated spline.
    }
    \vspace{-0.6cm}
    \label{fig:method_overview}
\end{figure*}
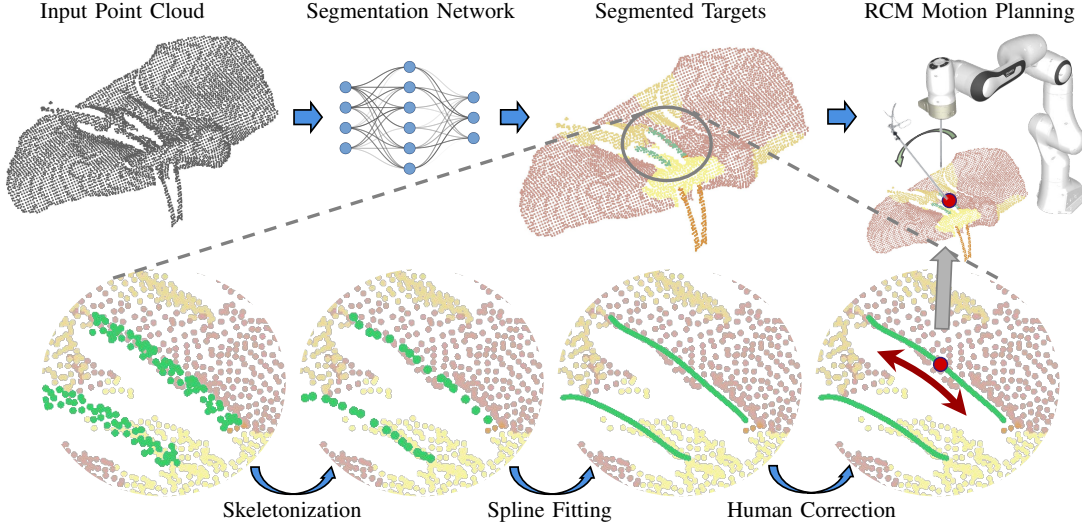

\glsreset{lc}

During \gls*{lc}, the cystic duct and cystic artery (the \textit{target structures}) must be transected (cut) before the gallbladder can be removed.
To prevent postoperative complications caused by bleeding or bile leaking into the abdominal cavity, these structures must be sealed, where best practice is to place 3 surgical clips on each vessel prior to cutting.
While guidelines for clip placement vary across surgical centers, the areas of the cystic duct and artery where clips may be applied are referred to as \textit{target regions}.

We develop a robotic system to autonomously position surgical clips (see \autoref{fig:method_overview}), while addressing several unique challenges.
First, the target positions must be predicted with sub-millimeter accuracy, otherwise the clip would not intersect with the target structure. %
However, the accuracy along the longitudinal axis of the target structures is less critical, provided the clip is a safe distance from the common bile duct.
The ordering of the six clips is also not strictly defined, allowing for flexibility in execution. %
Furthermore, the system must provide explicit plans that the surgeon can review to ensure safe operation. 
Finally, due to the absence of publicly available datasets for this task, and the overall data scarcity in the surgical domain, it must be highly data efficient.

While it is possible to directly predict the coordinates of the six targets with an end-to-end vision model, this method is presented as an ablation and fails to reach the required accuracy.
Instead, our approach leverages the accuracy of the point cloud camera by training a model to segment the target regions within the point cloud.
Each target region is fitted to a B-spline, from which three target points are selected by varying the spline parameter.
Compared to selecting a single point as the target, spline interpolation can be thought of as averaging over multiple points, resulting in a more robust estimation of the true surface. %
Using splines as targets also allows the system to predict an optimal angle of approach for the clip by computing the tangent to the spline at the target location.
During execution, the surgeon retains the ability to decide which clip to target next and adjust the target position along the spline as needed.
This approach shifts the model’s task from direct target prediction to a more well-defined segmentation problem, while allowing the surgeon to control task-specific degrees of freedom and make necessary corrections.

\paragraph{Experimental Setup}

A phantom scene is constructed containing a liver, a gallbladder, the cystic duct and artery, and fatty tissue (see \autoref{fig:real_setup}).
This scene represents the phase of the surgery after the surgeon has dissected the tissue covering the cystic duct and artery (Calot's triangle) and positively identified these structures.
The liver is based on an anonymized CT scan of a real patient at Heidelberg University Hospital and is 3D printed from red ABS at $1.15$X scale.
A gallbladder mold is printed based on a scan of the same patient from SLA material.
Several gallbladder phantoms are then produced by painting liquid latex rubber onto the mould and peeling off the dried material.
The cystic duct and artery are made of silicone tubing with outer diameter \SI{5}{\mm}.
For attaching the tubing to the gallbladder, angled \textit{rodlets} are 3D printed from SLA, where the angle determines the position and shape of the target structures.

Point clouds are taken with a 3D camera (Zivid One+ M, Zivid AS) which achieves a precision of \textless \SI{0.1}{\mm}.
A laparoscopic clip applier (model 30444LR, Karl Storz SE \& Co. KG) is mounted to a 7-DoF robot arm (Franka Panda, Franka Robotics GmbH) such that the axis of the tool is perpendicular to the axis of the end-effector.
A laparoscopic grasper (model 33356ON, Karl Storz SE \& Co. KG) is additionally used to grasp the neck of the gallbladder to apply tension, and is fixed in place during clipping.
A 3D-printed ring is used to visually mark the \gls*{rcm}, where in a surgery the trocar would be located.

\paragraph{Synthetic and Real Datasets}
\label{}

To address the data scarcity for this specialized task, we pre-train the model on synthetic data and fine-tune it using a small, manually labeled, real-world dataset.
To minimize the sim-to-real gap, we use point clouds without color features.

A synthetic dataset is created in Blender, an open-source 3D modeling and animation software.
Rather than attempt to replicate the setup in Blender exactly, we deliberately simplify the scene to demonstrate the robustness of our method across a significant sim-to-real gap.
We reuse the scans of the patient’s liver and gallbladder for the virtual scene. %
To introduce variability, random deformations are applied to the liver, gallbladder, and fatty tissue using a displacement field.
The pose of the gallbladder neck is randomized, with the target structures following B-spline curves.
Additionally, the diameters of the target structures are varied between \SI{2}{\mm} and \SI{4}{\mm}.
For simplicity, collisions between objects are ignored.

To ensure robustness to changes in the camera pose, we randomize the virtual camera with an incline from the range $[30\degree,70\degree]$ and distance from the scene from the range $[30,40]$\SI{}{\cm}.
Segmented depth images are rendered using the virtual camera and converted to point clouds, and each point is labeled with its class of origin: \textbf{Liver/Background}, \textbf{Gallbladder}, \textbf{Fatty Tissue}, \textbf{Cystic Duct}, or \textbf{Cystic Artery}.
Points on the cystic duct or cystic artery located within a distance of \SI{30}{\mm} from the middle of each structure are relabeled as \textbf{Target Duct} or \textbf{Target Artery}, respectively. 
Although we are only interested in the target regions, we show in our ablations that adding extra classes helps the network to learn a fine-grained understanding of the scene and promotes generalization.
We generate $128.000$ point clouds with roughly $4650$ points each.
The point clouds undergo preprocessing through voxel downsampling with a \SI{3}{\mm} grid. %

For the fine-tuning dataset, we record $60$ point clouds from the real experimental setup.
For each point cloud, we vary the liver pose, the distribution of fatty tissue, the grasping position of the gripper, and the gallbladder pose, while maintaining the same camera angle.
The point clouds are segmented manually in CloudCompare, an open-source software.
To minimize labeling effort, the real dataset only contains the \textbf{Background}, \textbf{Target Duct}, and \textbf{Target Artery} labels.
Each point cloud is pre-processed by cropping, removing statistical outliers, removing the plane of the table using RANSAC, and voxel downsampling with a \SI{3}{\mm} grid.

\paragraph{Pre-Training}

We train a neural network to segment point clouds into the seven classes defined above.
We choose the segmentation variant of PointTransformer~\cite{zhaoPointTransformer2021} because of its state-of-the-art performance on standard benchmark datasets.
The neural network is trained for $30$ epochs with a batch size of $48$ and an exponentially decaying learning rate of $\alpha(t) = \alpha_0 \cdot \beta^t$, where $\alpha_0=5 \cdot 10^{-4}$, $\beta=0.98$, and $t$ is the epoch.
The class weights in the cross-entropy loss are adjusted according to the frequency of each class in the dataset, to ensure there is no bias towards large classes.

On each forward pass, we apply data augmentation to the point clouds to increase the ability of the network to generalize.
Since we found the real scene to contain several occlusions due to the \gls*{rcm} ring or the grasper tool, we apply a novel form of data augmentation called \textit{RandomPatches} inspired by similar methods in the image domain~\cite{devries2017improved}.
For each point cloud, this process chooses a random number of patches to remove between $0$ and $30$, and a random patch size $k$ between $0$ and $200$ points.
The patch centers are sampled uniformly from the point cloud, and the patches are the $k$ nearest neighbors to the patch center.
The patches (including the patch centers) are removed from the point cloud.
Next, a random rotation from the range $[0\degree,90\degree]$ is applied in each of the three axes, followed by a random shear with magnitude from $[0,0.3]$, and a random stretch with magnitude from $[0,0.3]$.
Finally, we apply a novel augmentation method called \textit{VariableJitter}.
While adding random jitter to a point cloud tends to improve model robustness, standard jitter with a fixed scale always creates point clouds with the same expected noise, creating a challenging trade-off.
Using a larger jitter scale increases robustness (up to a certain point), but also creates a significant domain gap compared to point clouds recorded by the Zivid camera, which are highly precise and orderly.
To address this, \textit{VariableJitter} dynamically adjusts the jitter scale for each point cloud. 
It samples the jitter scale $d_i$ for the entire point cloud from the range $[0,d_\textrm{max}]$, where $d_\textrm{max} = 0.04$, and the displacement of each point is then sampled from $[0,d_i]$.
This ensures that some augmented point clouds maintain low noise, better matching the characteristics of the real data.

\paragraph{Fine-Tuning}

\begin{figure}[tb]
    \centering
    \input{figures/pivotized_instrument}
    \vspace{-0.1cm}
    \caption{
        In laparoscopic surgery, instruments are inserted into the abdomen through trocars, which imposes a pivotized motion constraint.
        This reduces the end effector's independent degrees of freedom to $3$ rotations (roll, pitch, \& yaw) and one translation (depth) around the remote center of motion.
        Illustration adapted from~\cite{scheiklLapGymOpenSource2023}.
    }
    \label{fig:pivotization}
    \vspace{-0.6cm}
\end{figure}
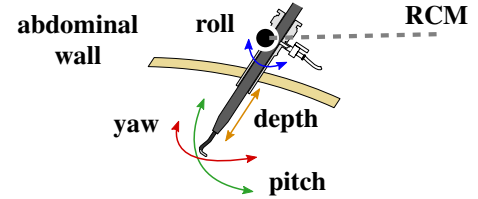

After pre-training, the neural network is fine-tuned on the real dataset for an additional $50$ epochs with a batch size of $8$ and a reduced initial learning rate of $\smash{\alpha_0 = 2.5 \cdot 10^{-4}}$.
In order not to overfit on the limited fine-tuning data, only the MLP head and the three upsampling layers (each composed of a TransitionUp layer and a transformer block) are optimized during fine-tuning.
Real point clouds undergo weaker data augmentation.
We use random rotations from the range $[0\degree,45\degree]$, random shear and random stretch from the range $[0,0.2]$, and \textit{VariableJitter} with maximum displacement $d_\textrm{max}=0.01$.

Since the fine-tuning dataset has fewer classes (three) than the neural network has outputs (seven), we investigate two methods for calculating the cross-entropy loss despite this limitation.
All classes except for the two target regions must be merged into a single background class. %
One approach is to combine the logits in probability space using \texttt{log-sum-exp}, which we call \textit{Probability-Merging}.
A simpler approach is to directly sum the logits before converting to probabilities, which we call \textit{Logit-Merging}.
Depending on the absolute scale of the logits, this can distort the loss landscape.
For example, with negative logit values, the probability of the merged class would be less than the probability of any of the individual classes.
Counterintuitively, we show in our ablations that \textit{Logit-Merging} results in the highest accuracy.

\paragraph{Target Localization}

The points belonging to each target region are skeletonized using Laplacian-Based Contraction~\cite{caoPointCloudSkeletons2010}.
Before interpolation, the points must be sorted along the major axis of the skeleton, the direction of which is calibrated based on the layout of the scene.
The sorted points are then fitted to a B-spline of degree $3$ using a smoothing factor of $5 \cdot 10^{-4}$.

A simple user interface allows a human operator to approve the computed target positions and adjust them if needed.
This allows the surgeon to adjust the desired distance between the two distal clips that remain in the body, or to correct minor errors resulting from the segmentation.
The UI also allows the operator to decide the order of the clips.

\paragraph{Metrics}

\begin{figure}[t]
    \centering
    \includegraphics[trim={0 2.55in 5.5in 0},clip,width=0.4\textwidth]{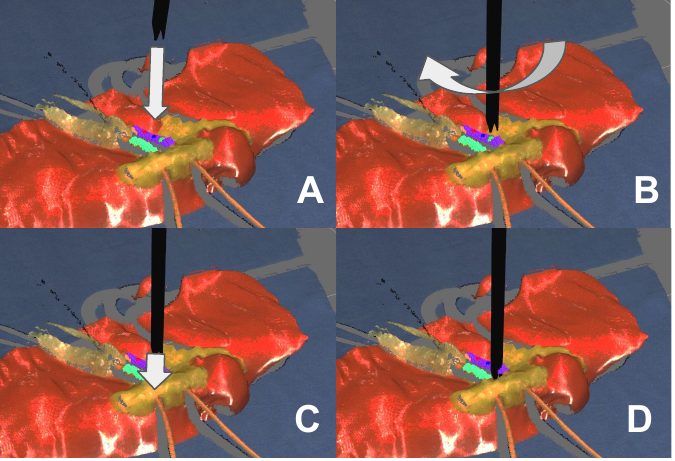}
    \caption{The clip is applied onto the target structure in three phases. A) navigate to within $\sim$\SI{2}{\cm} of the target, B) rotate about clipper such that insertion is not obstructed, and C) final insertion, placing clip around target structure.}
    \label{fig:path_planning}
    \vspace{-0.5cm}
\end{figure}

We report the accuracy of target localization using five metrics.
All metrics are computed across an unseen real-world test set collected during experiments on the real phantom, as described in Section \ref{sec:real_experiments}.

To measure segmentation performance, we report mIoU, the mean Intersection over Union across all classes.
We also report \textit{Target mIoU}, the mean Intersection over Union across only the two target classes, since they are the most relevant for target localization.
The \textit{Target mIoU} allows for a direct comparison between synthetic and real data, which is not possible with mIoU since they have different numbers of classes.
If too few points are segmented in a target region ($<15$ points), the target is considered not to have been found, and no spline is generated.
The \textit{Spline Found Rate} expresses the fraction of target regions for which a spline is successfully found.

To quantify the similarity of the predicted spline $S_\textrm{pred}$ to the reference spline $S_\textrm{ref}$, we discretize each and compute the asymmetrical chamfer distance between them.
First, we evaluate $\smash{N = M = 200}$ evenly-spaced points along each spline to create point clouds $P_\textrm{ref} = \{x_i \in \mathbb{R}^3 \}_{i = 1}^N$ and $P_\textrm{pred} = \{x_i \in \mathbb{R}^3 \}_{j = 1}^M$.
We remove 10\% from each end of the reference spline to focus on likely clip targets.
The asymmetrical chamfer distance between two point clouds is
\begin{equation}
    \textrm{chamfer}(P_\mathrm{ref}, P_\mathrm{pred}) = \frac{1}{N} \sum_{i = 1}^N \| x_i - \mathcal{NN}(x_i, P_\mathrm{pred}) \| ,
\end{equation}
where $\mathcal{NN}(x, P) = \mathrm{argmin}_{x' \in P} \|x - x'\|$ is the nearest neighbor to $x$ in the point cloud $P$.
We choose the asymmetrical chamfer distance because it quantifies the deviation of the reference spline from the predicted spline, while not punishing the model if parts of the predicted spline are far from the reference.
In practice, this measures how close a human operator could get to a desired target point on the reference spline while being constrained to the predicted spline.

\textit{Success Rate} reports the fraction of predicted splines with a chamfer distance below a threshold of \SI{0.75}{\mm}.
This threshold is half the clearance between target structure (\SI{5}{\mm}) and the clip (\SI{6.5}{\mm}), and does not take into account errors in camera calibration, motion planning, or robot control.
The success rate is averaged across all target regions, where if no spline is found, the target is considered unsuccessful.

\paragraph{Motion Planning}

\glsreset{rcm}

First, we estimate the static transformation required to transform the target positions from camera coordinates into the robot’s reference frame.
This is obtained via least-squares regression using 12 pairs of corresponding points from both coordinate systems.
These correspondences are found by positioning the clip applier at the corners of a calibration checkerboard visible in the camera frame.

A key aspect of robot-assisted minimally invasive surgery is to satisfy the remote center of motion constraint about the trocar, which reduces the number of independently controllable DoFs from 6 to 4 (shown in \autoref{fig:pivotization}) .
Therefore, we opt to use a task space where the degrees of freedom are the same as those of a laparoscopic instrument inside a trocar: three rotations, roll, pitch, yaw; and one translation along the instrument axis, depth.
We plan the motion to the target positions in task space and compute the corresponding trajectory in Cartesian space. %
The resulting trajectory naturally satisfies the constraints of pivotized motion around a remote center of motion, and is densely sampled so that it can be used directly for Cartesian control of the robot.

In order to ensure the clip is properly aligned when it approaches the target structure, the clipping motion is divided into 3 stages, approach, align, and insert, as shown in \autoref{fig:path_planning}.
A single waypoint a distance of \SI{2.3}{\cm} (the length of the clip applier) from the target ensures that the alignment is step finished and the clip is correctly rotated before the insertion step begins.
Each planned motion is verified by the human operator before execution.
After insertion, the clipping itself is actuated manually by the human operator.

\section{RESULTS AND DISCUSSION}
\label{sec:real_experiments}
\paragraph{Clipping on Real Phantom}

\begin{figure}[tb]
    \centering
    \includegraphics[trim={0 0 2.0in 0},clip,width=0.8\linewidth]{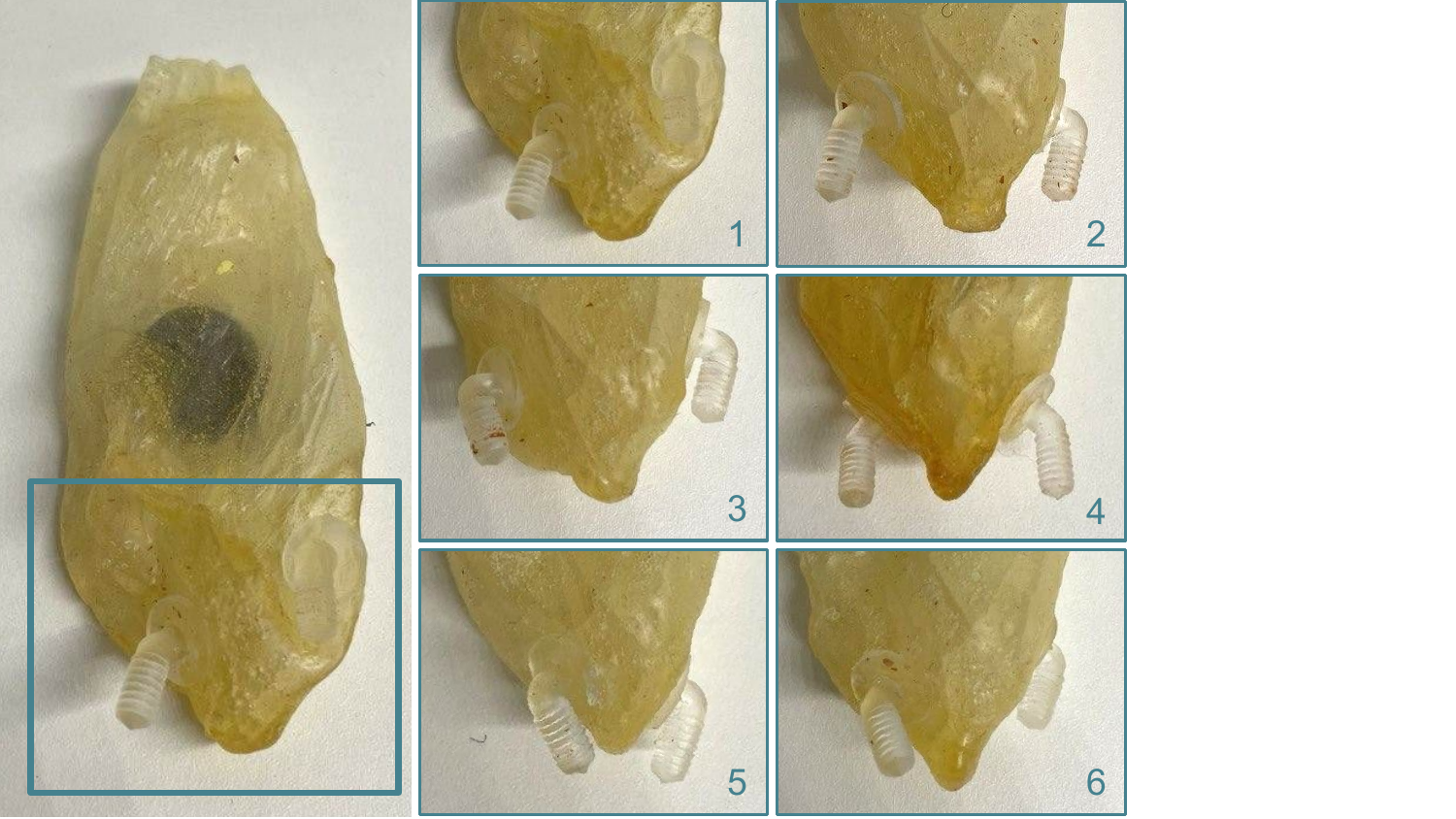}
    \caption{
        Different gallbladder phantoms used for real world experiments.
        The cystic duct and artery are mounted at unique positions and angles on each gallbladder through the use of the clear plastic rodlets highlighted by the teal box.
        The numbered insets show the variation between gallbladders.
    }
    \label{fig:gallbladders}
    \vspace{-0.5cm}
\end{figure}

We demonstrate our method for autonomous clip positioning on the real-world setup described above. %
A medical student who has completed a laparoscopic training curriculum acts as the operator.
We apply six clips each to six novel gallbladder phantoms with different morphologies not seen in the fine-tuning data (see \autoref{fig:gallbladders}).
In particular, the angles of the target structures to the gallbladder are varied by using rodlets with different angles, resulting in novel point clouds not seen during training.

Out of the $36$ clips, the operator identified possibly incorrect segmentations in $5$ cases.
In each of these instances, the operator repeated point cloud acquisition and model inference, which then produced a correct segmentation and target localization, partly due to the stochastic nature of model inference.
This implies a success rate of $\frac{36}{36 + 5} = 87.8\%$ for target localization, though this should be considered a lower bound.
It is possible that some of the incorrect segmentations could have still led to successful clipping, especially if the operator had adjusted the target points.
In our experiments, the operator never adjusted target points from their default positions.
For cases with successful segmentations, we report a $100$\% success rate for autonomous clip positioning, with $36/36$ clips applied correctly.
Given the dimensions of the clips and the phantom, this implies a total error of \textless\SI{1}{\mm} across target localization, camera calibration, motion planning, and robot control.

During testing, we record $2$ point clouds taken before each clip, which are then manually labeled to create reference segmentations.
The reference target regions are skeletonized and fitted to splines to create a test dataset of $72$ point clouds.
This test dataset is used to evaluate post-hoc how several ablations of our method would have performed during real world experiments, allowing us to evaluate many more variations than would be feasible on the real robot.
These \textit{in silico} results are reported as a mean over $20$ random seeds for fine-tuning, and no data augmentation is applied during evaluation.
When evaluated on this dataset, our method achieves an mIoU of $0.744$, a target mIoU of $0.619$, a chamfer distance of \SI{0.3}{\mm}, and a success rate of $94.9\%$, which is indeed slightly higher than the observed success rate.

\paragraph{Target Point Regression}

To validate our segmentation-based approach, we present an ablation where the six target positions are directly predicted by the network.
Reference target positions are computed for each point cloud in the synthetic and real datasets by evaluating each reference spline at points $35\%$, $50\%$ and $65\%$.
We modify the classification variant of PointTransformer to regress $18$ values, representing the xyz coordinates of the six target points.
The symmetric chamfer distance is used as the loss function for both pre-training and fine-tuning, since the regressed target points are interchangeable.
Because this network does not have upsampling layers, we opt to unfreeze the entire network for fine-tuning.
We fine-tune for $100$ epochs instead of $50$, since convergence takes longer.

We report a mean distance between the reference and predicted targets of \SI[separate-uncertainty=true]{5.7(5.2)}{\mm}.
Applying a wider threshold of \SI{1}{\mm} for success, this results in a success rate of $1$\%.
However, this evaluation is simplified, since predicting only target positions would leave the angle of approach undefined.
Adding pose prediction to the network is possible but would likely further reduce accuracy.
Therefore, target point regression cannot achieve the required accuracy for this task.

\paragraph{Pre-Training Dataset}

\begin{table}
    \caption{Ablation Pre-Training \& Fine-Tuning}
    \label{tab:ablate_classes}
    \centering
    \begin{threeparttable}
        \begin{tabular}{c p{0.65cm} p{0.65cm} p{1.0cm} p{0.8cm}}
            \toprule
             & mIoU & Target mIoU & Chamfer Distance & Success Rate \\
            \midrule
            Pretrained only$^\dagger$ & 0.884 & 0.809 & - & - \\
            \textbf{Logit-Merging} & \textbf{0.744} & \textbf{0.619} & \textbf{\SI{0.3}{\textbf{\mm}}} & \textbf{94.9\%} \\
            Probability-Merging & 0.481 & 0.239 & \SI{1.3}{\mm} & 41.4\% \\
            No Additional Classes & 0.534 & 0.313 & \SI{0.8}{\mm} & 63.3\% \\
            No Pre-Training & 0.346 & 0.096 & \SI{12.0}{\mm} & 7.6\% \\
            \bottomrule
        \end{tabular}
        \begin{tablenotes}
            \footnotesize
            \item $^\dagger$ On a held-out synthetic test dataset of 12,800 point clouds
        \end{tablenotes}
    \end{threeparttable}
    \vspace{-0.5cm}
\end{table}

We investigate two methods for merging classes when fine-tuning on a dataset with fewer classes than the model has outputs, \textit{Logit-Merging} and \textit{Probability-Merging}. %
We compare this against pre-training with no additional classes, such that the network only predicts the classes \textit{Background}, \textit{Target Duct}, and \textit{Target Artery}, while the remaining classes are all merged into the \textit{Liver/Background} class.
We additionally compare against no pre-training at all.
To keep the comparison fair, the network still predicts $7$ classes, and we merge additional classes using \textit{Logit-Merging}.
We also train for $100$ epochs instead of $50$, since it takes longer for the network to converge.

The results are summarized in \autoref{tab:ablate_classes}.
\textit{Logit-Merging} far outperforms \textit{Probability-Merging}, mainly due to a much lower chamfer distance (\SI{0.3}{\mm} vs. \SI{1.3}{\mm}).
Pre-training without additional classes results in a chamfer distance of \SI{0.8}{\mm}, which is better than \textit{Probability-Merging}, although it still only results in a success rate of $63.3\%$.
\textit{Logit-Merging} shows a meaningfully higher Target mIoU of $0.619$ compared to \textit{Probability-Merging} ($0.239$) or training without additional classes ($0.313$), which is reflected in its high success rate.
Training only on the real dataset without any pre-training, in contrast, achieves only a 7.6\% success rate, despite training for $100$ epochs instead of $50$.

\paragraph{Data Augmentation}

\begin{table}
    \centering
    \caption{Ablation of Data Augmentation Methods}
    \resizebox{0.49\textwidth}{!}{%
    \begin{tabular}{wc{2.1cm} p{0.65cm} p{0.65cm} p{0.8cm} p{1.0cm} p{0.8cm}}
        \toprule
         & mIoU & Target mIoU & Spline Found Rate & Chamfer Distance & Success Rate \\
        \midrule
        \textbf{All} & \textbf{0.744} & \textbf{0.619} & \textbf{99.6\%} & \textbf{\SI{0.3}{\textbf{\mm}}} & \textbf{94.9\%} \\
        w/o Shear/Stretch & 0.727 & 0.593 & 97.3\% & \SI{0.6}{\mm} & 81.1\% \\
        w/o VariableJitter & 0.631 & 0.451 & 86.6\% & \SI{1.0}{\mm} & 62.5\% \\
        w/o RandomPatches & 0.635 & 0.533 & 89.0\% & \SI{1.0}{\mm} & 58.8\% \\
        w/o Jitter & 0.396 & 0.099 & 11.4\% & \SI{1.7}{\mm} & 4.9\% \\
        w/o Rotations & 0.330 & 0.0 & 0.0\% & N/A & 0.0\% \\
        w/o Augmentation & 0.467 & 0.205 & 30.0\% & \SI{1.9}{\mm} & 16.8\% \\
        \midrule
        \makecell[c]{PointWOLF instead\\of VariableJitter} & 0.632 & 0.452 & 76.6\% & \SI{1.2}{\mm} & 61.3\% \\
        \makecell[c]{PointWOLF instead\\of RandomPatches} & 0.686 & 0.531 & 78.6\% & \SI{1.2}{\mm} & 42.4\% \\
        \bottomrule
    \end{tabular}
    }
    \label{tab:ablate_data-augmentation}
    \vspace{-0.5cm}
\end{table}

Next, we show that data augmentation is essential for allowing the network to generalize across the distribution gap between the real and synthetic data.
Since data augmentation can be applied independently during fine-tuning and pre-training, the space of augmentation combinations is large.
We choose to disable individual augmentation steps in both pre-training and fine-tuning simultaneously, testing the overall contribution of each method.
Additionally, we validate our novel augmentation methods \textit{VariableJitter} and \textit{RandomPatches} by replacing them with PointWOLF~\cite{kimPointCloudAugmentation2021}, a general-purpose augmentation method that applies a random deformation field to each point cloud.
Note that \textit{RandomPatches} is never enabled during fine-tuning.
The ablation ``w/o VariableJitter" substitutes standard random jitter instead of \textit{VariableJitter}, whereas ``w/o Jitter" disables all forms of jitter completely.

The results are summarized in \autoref{tab:ablate_data-augmentation}.
Replacing \textit{VariableJitter} with standard jitter decreases the success rate to 62.5\%, making this an essential element of the overall system success.
We hypothesize that this was particularly important due to the extremely low noise of the Zivid camera, which produces point clouds that look very dissimilar to typical jittered point clouds.
Removing \textit{RandomPatches} reduces the success rate to only 58.8\%, likely because the model is less robust to missing regions of the point cloud caused by occlusions.
PointWOLF does not address either of these challenges, and is therefore not able to achieve the same level of success.

Removing either random jitter or random rotations, while keeping all other augmentation methods unchanged, renders the model completely unable to learn the fine-tuning data.
During pre-training, these models achieve Target mIoUs of 0.889 and 0.736, respectively, on the training data; however, during fine-tuning, their Target mIoUs on the training data drop to 0.063 and 0.0.
This indicates an inability to generalize across the sim-to-real gap.
Notably, these configurations perform even worse than training without data augmentation at all, which yields a 16.8\% success rate.

\paragraph{Network Architecture}

\begin{table}
    \centering
    \caption{Ablation of Network Architecture}
    \begin{tabular}{c p{0.65cm} p{0.65cm} p{1.0cm} p{0.8cm}}
        \toprule
         & mIoU & Target mIoU & Chamfer Distance & Success Rate \\
        \midrule
        PointNet++~\cite{qiPointNetDeepHierarchical2017} & 0.489 & 0.249 & \SI{1.1}{\mm} & 40.3\% \\
        Stratified Transformer~\cite{lai2022stratified} & \textbf{0.796} & \textbf{0.696} & \SI{0.4}{\mm} & 83.1\% \\
        \textbf{PointTransformer}~\cite{zhaoPointTransformer2021} & 0.744 & 0.619 & \textbf{\SI{0.3}{\textbf{\mm}}} & \textbf{94.9\%} \\
        \bottomrule
    \end{tabular}
    \label{tab:ablate_network}
    \vspace{-0.5cm}
\end{table}

We investigate the effectiveness of several common neural network architectures for point cloud segmentation.
PointNet++\cite{qiPointNetDeepHierarchical2017} introduced hierarchical processing of point clouds.
Stratified Transformer~\cite{lai2022stratified} is able to capture long-range contexts by sampling both nearby and distant points as keys for each query point.
Stratified Transformer appears to perform better if the entire network is fine-tuned, so this is the configuration we report here.

The results are summarized in Table \ref{tab:ablate_network}.
PointNet++ is hindered by poor segmentation performance (e.g. Target mIoU of 0.249), leading to low success. %
Although Stratified Transformer performs well under the mIoU and Target mIoU metrics, which indicate good segmentation performance, its success rate is lower than Point Transformer.
We observe that this is due to a lower Spline Found Rate (91.1\% vs. 99.7\%), which is not captured by the mIoU, in which empty classes are ignored when taking the mean.
Overall, we hypothesize that PointTransformer may be better at generalization when the fine-tuning dataset is extremely small.

\paragraph{Fine-Tuning}

Since PointTransformer consists of downsampling and upsampling layers in a U-Net-like configuration, we investigate different configurations of freezing or training network weights during fine-tuning.
We test unfreezing $1$, $2$, or all $3$ of the upsampling layers, as well as fine-tuning the entire network.
We find that fine-tuning all $3$ upsampling layers is substantially better than $1$ or $2$ layers, which achieve success of $2.2\%$ and $8.1\%$, respectively.
Fine-tuning the entire network results in a chamfer distance of \SI{1.9}{\mm} and the success rate drops to $59.3\%$.

\paragraph{Size of Synthetic Dataset}

Finally, in \autoref{tab:ablate_data-size}, we describe the effect of reducing the amount of synthetic data for pre-training.
Greater amounts of synthetic data are indeed beneficial, but the returns are diminishing.
There is a small but meaningful improvement between  $64,000$ and $128,000$ samples.
Using only $12,800$ samples results in noticeably worse performance, with only an $60.7\%$ success rate.

\begin{table}
    \centering
    \caption{Ablation of Size of Synthetic Pre-Training Dataset}
    \begin{tabular}{c p{0.65cm} p{0.65cm} p{1.0cm} p{0.8cm}}
        \toprule
        Dataset Size & mIoU & Target mIoU & Chamfer Distance & Success Rate \\
        \midrule
        10\% & 0.714 & 0.574 & \SI{1.3}{\mm} & 60.7\% \\
        50\% & 0.732 & 0.601 & \SI{0.5}{\mm} & 88.2\% \\
        \textbf{100\%} & \textbf{0.744} & \textbf{0.619} & \textbf{\SI{0.3}{\textbf{\mm}}} & \textbf{94.9\%} \\
        \bottomrule
    \end{tabular}
    \label{tab:ablate_data-size}
    \vspace{-0.5cm}
\end{table}

\section{CONCLUSIONS}

We present the first system to achieve autonomous clip positioning in laparoscopic surgery on a physical phantom using point clouds.
Our segmentation-based approach leverages the precision of the 3D camera, making it possible to localize targets with the required accuracy of \textless\SI{1}{\mm}, while retaining interpretability.
We use motion planning to ensure that the \gls*{rcm} constraint imposed by the trocar is satisfied and that the system remains interpretable.
We demonstrate our method with a real robot on six gallbladder phantoms and report success rates of $95$\% for target localization and $100\%$ for autonomous clip positioning.
We validate our algorithmic contribution post-hoc on the real robot data: pre-training with additional semantic classes, and \textit{VariableJitter} and \textit{RandomPatches} for data augmentation.
Our contributions can be applied to other robotic applications where accuracy and interpretability are core challenges, and represent a significant step towards translating the promise of machine learning for surgical robotics to the clinical reality.

\paragraph{Limitations and Future Work}

While the proposed robotic system demonstrates high accuracy and interpretability, certain limitations remain.
Notably, the Zivid camera, selected for its exceptional accuracy, is not suitable for clinical deployment due to its physical size.
Initial experiments with a more compact and widely available stereo camera (ZED Mini, StereoLabs Inc.) revealed that its limited accuracy was insufficient for our method, which currently cannot compensate for systematic offsets in observed point positions.
Future work could explore integrating 3D reconstruction methods~\cite{peterStereoReconstructionMicroscopic2025} to infer true point positions from noisy observations, potentially reducing the accuracy required of the 3D camera.
Additionally, the motion planner does not take collision avoidance into account or respond to changes in the environment, so the human operator is responsible for verifying the safety of the planned motions and initiating re-planning when necessary.

\addtolength{\textheight}{-12cm}   %

{\small
\bibliographystyle{IEEEtran}
\bibliography{output.bbl}
}

\end{document}

%% file: figures/overview.tex
\tikzset{every picture/.style={line width=0.75pt}} %

\includegraphics{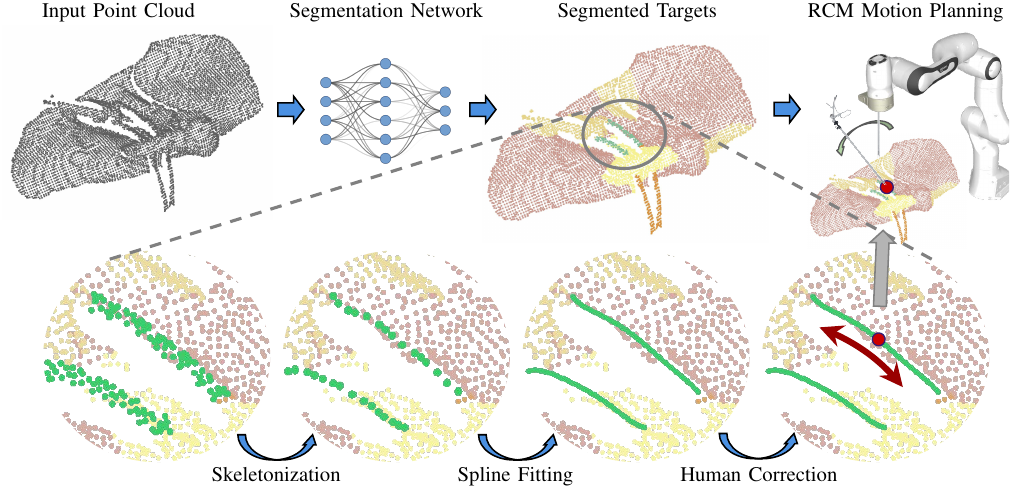}

%% file: figures/pivotized_instrument.tex
\includegraphics{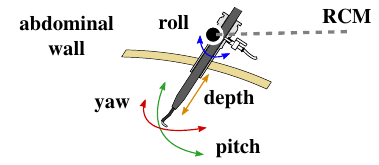}